**Research on Brain Tumor Classification Method Based on Improved ResNet34 Network**

Yufeng Li[*a], Wenchao Zhao[b], Bo Dang[c], Weimin Wang[d]

[*a] University of Southampton, Southampton, UK, liyufeng0913@gmail.com;

[b]University of Science and Technology of China, Anhui, China, ywzhaohong@gmail.com;

[c]Computer Science, San Francisco Bay University Fremont, CA, US, dangdaxia@gmail.com;

[d]The Hong Kong University of Science and Technology, Hongkong, China, wangwaynemin@gmail.com;

**Abstract**: Previously, image interpretation in radiology relied heavily on manual methods. However, manual classification of brain tumor medical images is time-consuming and labor-intensive. Even with shallow convolutional neural network models, the accuracy is not ideal. To improve the efficiency and accuracy of brain tumor image classification, this paper proposes a brain tumor classification model based on an improved ResNet34 network. This model uses the ResNet34 residual network as the backbone network and incorporates multi-scale feature extraction. It uses a multi-scale input module as the first layer of the ResNet34 network and an Inception v2 module as the residual downsampling layer. Furthermore, a channel attention mechanism module assigns different weights to different channels of the image from a channel domain perspective, obtaining more important feature information. The results after a five-fold crossover experiment show that the average classification accuracy of the improved network model is approximately 98.8%, which is not only 1% higher than ResNet34, but also only 80% of the number of parameters of the original model. Therefore, the improved network model not only improves accuracy but also reduces clutter, achieving a classification effect with fewer parameters and higher accuracy.

# 1.Introduction

With the rapid development of modern society and changes in lifestyle, people's health problems are becoming increasingly prominent. Among them, brain tumors, as a serious disease, are gradually becoming a health hazard troubling society. Brain tumors refer to a type of tumor that forms in the human brain tissue. They can affect a person's cognition, behavior, perception, and motor abilities, and their mortality rate accounts for 2.4% of the human tumor incidence rate[1]. There are many types of brain tumors, the most common of which are gliomas, meningiomas, and pituitary tumors[2]. The most commonly used method for diagnosing brain tumors in clinical practice is magnetic resonance imaging (MRI)[3]. MRI has multi-sequence and multi-type imaging capabilities, and can clearly display the soft tissue structure of the human body, providing richer imaging information for clarifying the nature of the lesion. Therefore, how to efficiently and accurately determine whether an MRI image contains a tumor and what type of tumor it contains is an urgent clinical problem to be solved.

Early MRI imaging relied on manual visual examination of brain MRI images. Due to the large data volume and the varying types of brain tumors, this method was time-consuming and prone to human error, making computer-aided diagnosis necessary[4]. In recent years, convolutional neural networks (CNNs) have achieved great success in many fields, such as image processing, speech recognition, natural language processing, object tracking, and medical diagnosis[4]. Along with this research wave, more and more scholars have introduced CNNs into the field of medical imaging in recent years. Hossain et al.[4] studied brain tumor detection using both traditional classifiers and deep learning-based CNNs. Their results showed that the detection performance using CNNs was significantly better than that of traditional classifiers. Abiwinanda[6] also applied CNNs to brain tumor classification, achieving a three-classification task of glioma, meningioma, and pituitary adenoma using a simple convolutional network. Although the accuracy was slightly lower, it was still comparable to the accuracy of traditional algorithms based on region preprocessing. Vankdothu et al.[7] combined CNNs with Long Short-Term Memory (LSTM) to construct a novel convolutional network, achieving a 92% accuracy rate in classifying gliomas, meningiomas, pituitary adenomas, and tumor-free images on the Kaggle brain tumor dataset. Ayadi et al.[8] constructed an 11-layer convolutional CNN to classify brain tumor images from the Figure dataset into three categories, achieving 94.74% accuracy. These models are shallow CNNs and do not yet reach the level of deep CNNs. Khan et al.[9] used a simple CNN model to classify benign and malignant brain tumors with 100% accuracy, but due to the small sample size (only 253 images), it was difficult to overcome the impact of the small sample size on the model's generalization performance.

As CNNs are applied to various fields, researchers have found that with increasing convolutional layer depth, gradient vanishing or exploding phenomena occur, leading to a significant drop in accuracy. Residual networks can mitigate this phenomenon. Kumar et al.[10] used residual networks and global average pooling to perform three-class classification on a brain tumor dataset, replacing fully connected layers with global average pooling to alleviate overfitting and achieving an accuracy of 97.48%. ResNet[11] residual networks can train very deep neural networks, effectively mitigating gradient vanishing and exploding during training, accelerating network convergence, and improving the model's expressive power and performance. The residual connections preserve original features, making network learning smoother and more stable, further improving the model's accuracy and generalization ability. However, ResNet extracts features at a single scale, without integrating features extracted from convolutional kernels at multiple scales. In CNNs, large convolutional kernels have a larger effective receptive field and higher shape bias, while small convolutional kernels focus more on texture bias[12]. Therefore, combining convolutional kernels of different sizes to process input data at multiple scales can appropriately improve model performance. Diaz et al.[13] applied a multi-scale approach to deep CNNs, performing a three-class classification on 3064 MRI images from a publicly available brain tumor dataset, achieving a classification accuracy of 97.3%. Some researchers have incorporated multi-scale concepts into residual networks[14], giving residual blocks multiple branches to control the number of channels, ensuring that the information received and features extracted differ in each convolution process. Furthermore, when processing input data, attention mechanisms can filter high-value information from a large amount of data, giving more attention to key information, enabling the model to better represent key features, while also reducing sensitivity to noise and redundant features, and improving the model's generalization ability.

To address the problems of time-consuming and labor-intensive traditional manual classification of brain tumor medical images, low classification accuracy of shallow CNN models, and gradient vanishing issues in stacked deep networks, this paper combines the advantages of multi-scale concepts and attention mechanisms to study a deep residual network that can alleviate the gradient vanishing problem. The main research aspects are: 1) Using a ResNet34 residual network as the backbone network, where the residual modules can effectively alleviate the gradient vanishing problem in deep networks; 2) Introducing multi-scale feature extraction into the backbone network, performing multi-scale feature extraction and concatenation on the first convolutional layer and downsampling layer of the backbone network to enrich the extracted feature information and ensure full utilization of input features during downsampling; 3) Finally, adding an SE channel attention mechanism module to the network to improve the representation ability of important features. Through these studies, a new network is proposed to improve the accuracy of brain tumor classification.

# 2. Network Structure

## 2.1 ResNet34 Residual Structure

ResNet34 is a specific implementation of the ResNet series of networks. Its main architecture includes convolutional layers, pooling layers, residual blocks, and fully connected layers. In ResNet, the residual block is one of its core ideas, and its shortcut connection branches allow shallow features to be directly mapped to deep network layers. This shortcut connection design helps alleviate the accuracy decline problem that occurs in deep neural networks as the number of layers increases. The structure of the residual block is shown in Figure 1. Its core idea is to achieve residual learning through cross-layer connections, which makes it easier for the network to learn residual information during training, thereby improving model performance and convergence speed. The innovation of this architecture has greatly promoted the development of deep convolutional neural networks, enabling us to build deeper neural networks without being constrained by performance degradation.

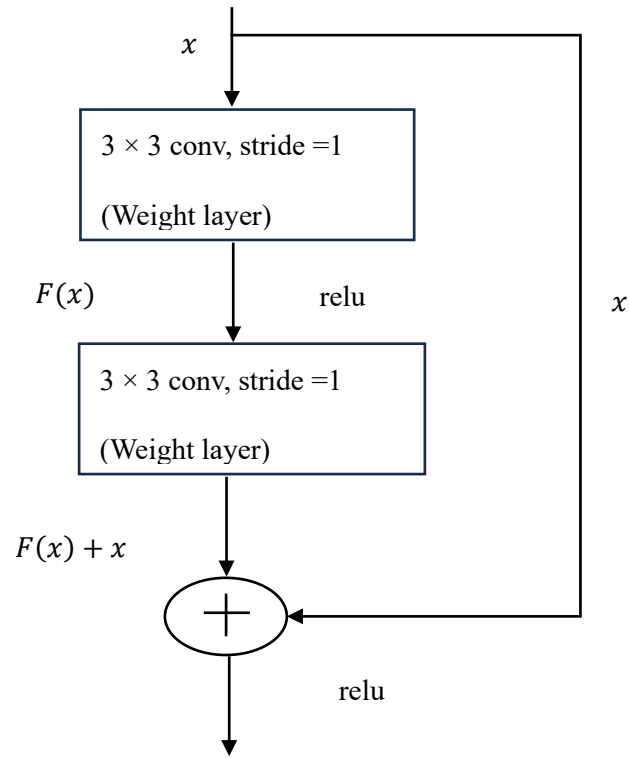

**Figure 1. Residual block structure diagram**

In Figure 1, the right branch $x$ represents the identity mapping of the residual structure, indicating that pixels are added one by one. The formula is as follows:

$$H(x) = F(x) + x \quad (1)$$

In the formula, $H(x)$ is the bottom layer mapping, $F$ represents the computation process through two convolutional layers (including ReLU activation operation), and $x$ is the input.

The residual mapping $F(x) = H(x) - x$, when the network reaches relative saturation, that is, when F(α) is small enough, the output $H(x)$ will approximately become α, which becomes the identity mapping function $H(x) = x$. The identity mapping ensures that the network performance will not decrease, so that the network can learn new features based on the input features. Moreover, the identity mapping will not add extra parameters and computation to the network, but can speed up the training speed of the model and optimize the training effect[15]. Therefore, as the network depth increases, the gradient will not disappear in backpropagation, which makes the residual network widely used in many fields.

## 2.2 Improved ResNet34 Network Model

To improve the ResNet34 network and overcome some of its limitations, this study optimized ResNet34. The improved network model still uses ResNet34 as the backbone network structure, but the architecture has been adjusted. First, the first 7×7 convolutional layer of ResNet34 is replaced with a multi-scale input module. This module can use convolutional kernels of various sizes to extract features from the image, thereby improving the richness and diversity of feature information. Second, the residual downsampling module in ResNet34 is replaced with the downsampling module proposed in this paper. This novel downsampling module not only introduces the multi-scale concept but also makes full use of all input information, further enhancing the model's performance. Finally, to further improve the network's performance, an SE attention mechanism is introduced in each residual block, which helps the network to learn and utilize key features more effectively. The network structure is shown in Figure 2.

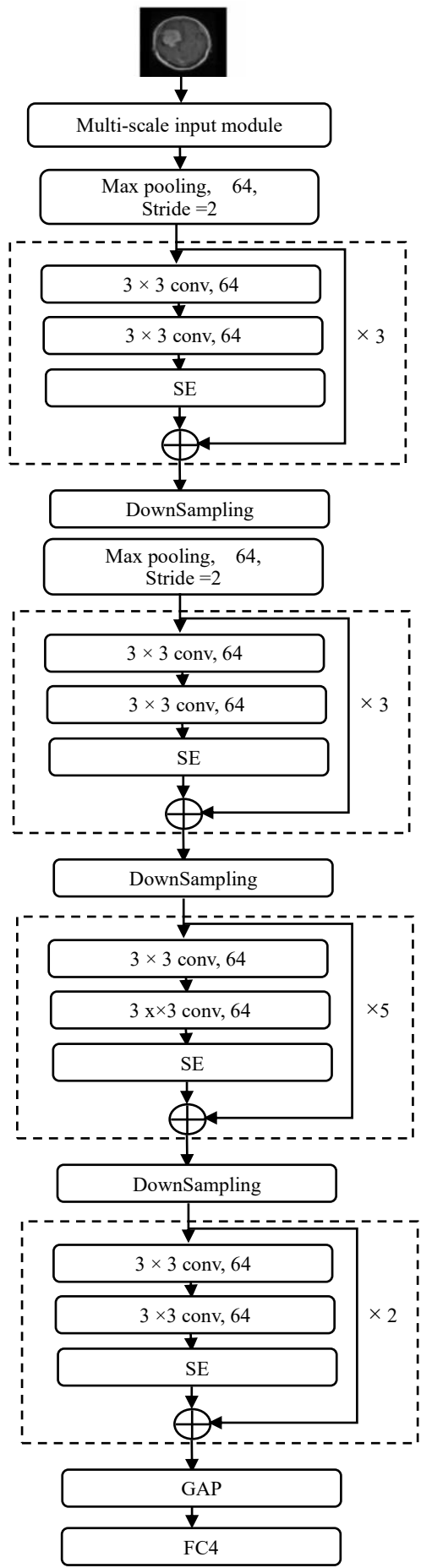

Figure 2　Improved ResNet34 network architecture diagram

## 2.3 Downsampling Module

Downsampling is an operation used to reduce the size or resolution of an image or signal, thereby reducing the computational cost of the model. In the ResNet34 network, in addition to using a max-pooling downsampling layer with a kernel size of 3×3 and a stride of 2, a residual-based downsampling module is also included, as shown in Figure 3. It is important to note that the shortcut connection of the downsampling module in Figure 3 is a convolutional layer with a kernel size of 1×1 and a stride of 2. This means that the input features passing through this convolutional layer will experience information loss, which may include some important information. To address this issue, combining the ideas of multi-scale and the Inception module, this paper proposes using the Inception v2 module as the downsampling module, changing the original 1×1 convolutional kernel branch to a 3×3 convolutional kernel. The network structure is shown in Figure 4. In Figure 4, `channels` represents the number of channels, `H` and `W` represent the height and width of the input feature map, respectively, and `stride` represents the stride of the convolution operation. `stride=2` indicates a downsampling operation, reducing the size of the input feature map to half its original size. This downsampling module has four branches. After each branch completes its operation, channels are concatenated. The final output channel count of each branch is the same, and the concatenated channel count is twice the input channel count.

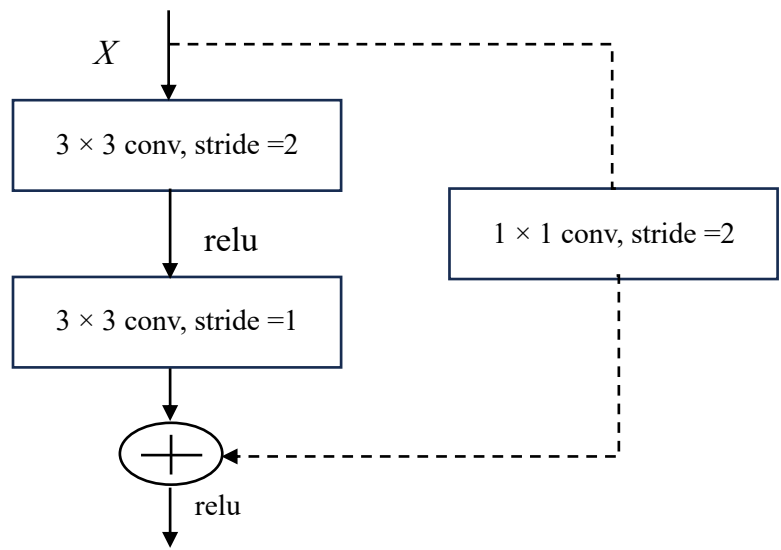

Figure 3 Differential block downsampling module

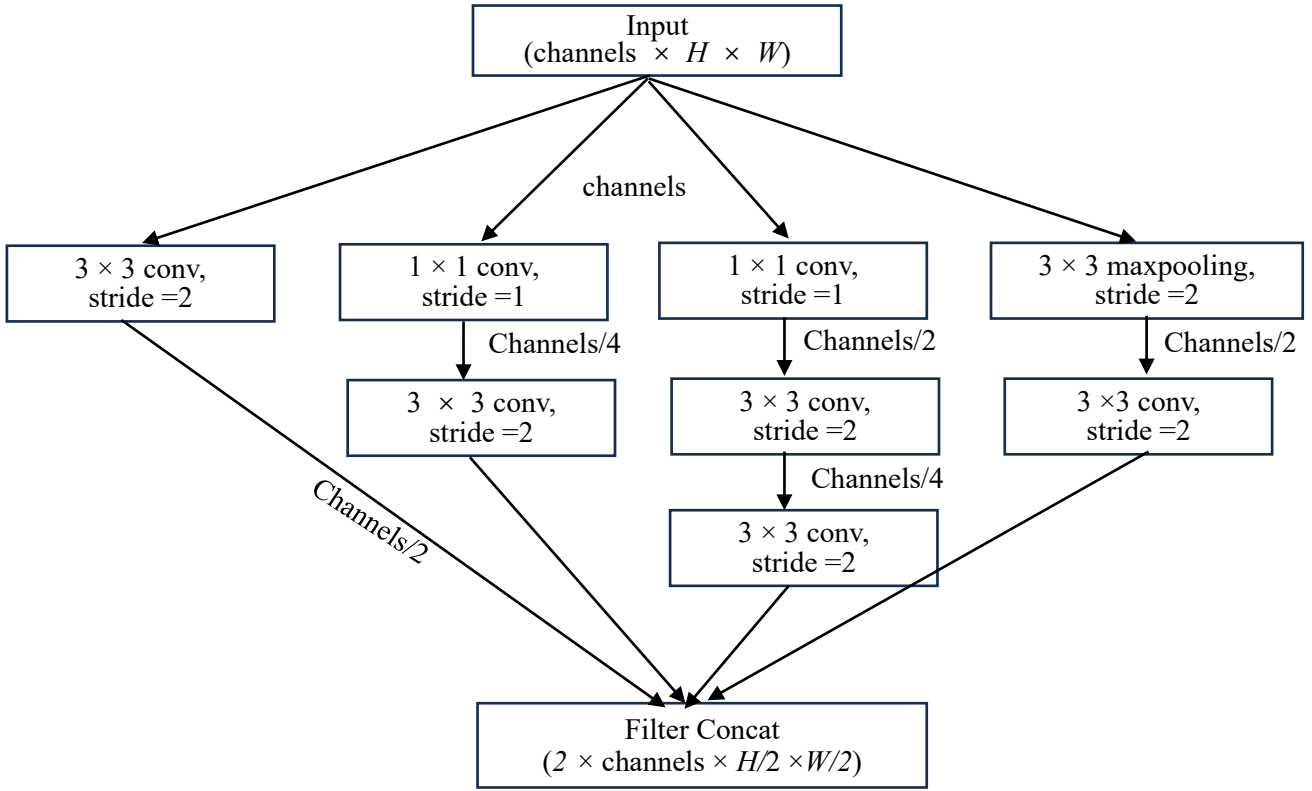

Figure 4 Inception v2 downsampling module

## 2.4 Multi-Scale Input Feature Extraction Module

Different sizes of convolutional kernels exhibit differences in feature extraction. Large convolutional kernels primarily focus on shape information, while small convolutional kernels emphasize capturing texture details[12]. The core idea of multi-scale feature extraction is to extract features separately using convolutional kernels of different sizes, and then add or concatenate their outputs to obtain richer feature information. The multi-scale input feature extraction module used in this paper first extracts features from the input data through four convolutional layers with kernel sizes of 3×3, 5×5, 7×7, and 11×11. Then, these four outputs are added pairwise to generate two intermediate outputs. Next, these two intermediate outputs are passed through two convolutional layers with kernel sizes of 3×3 and 5×5, respectively, and finally merged by channel concatenation. This multi-scale feature extraction strategy helps to capture various feature information in the image more comprehensively. The specific structure is shown in Figure 5, where 32 and 64 represent the number of output channels, respectively.

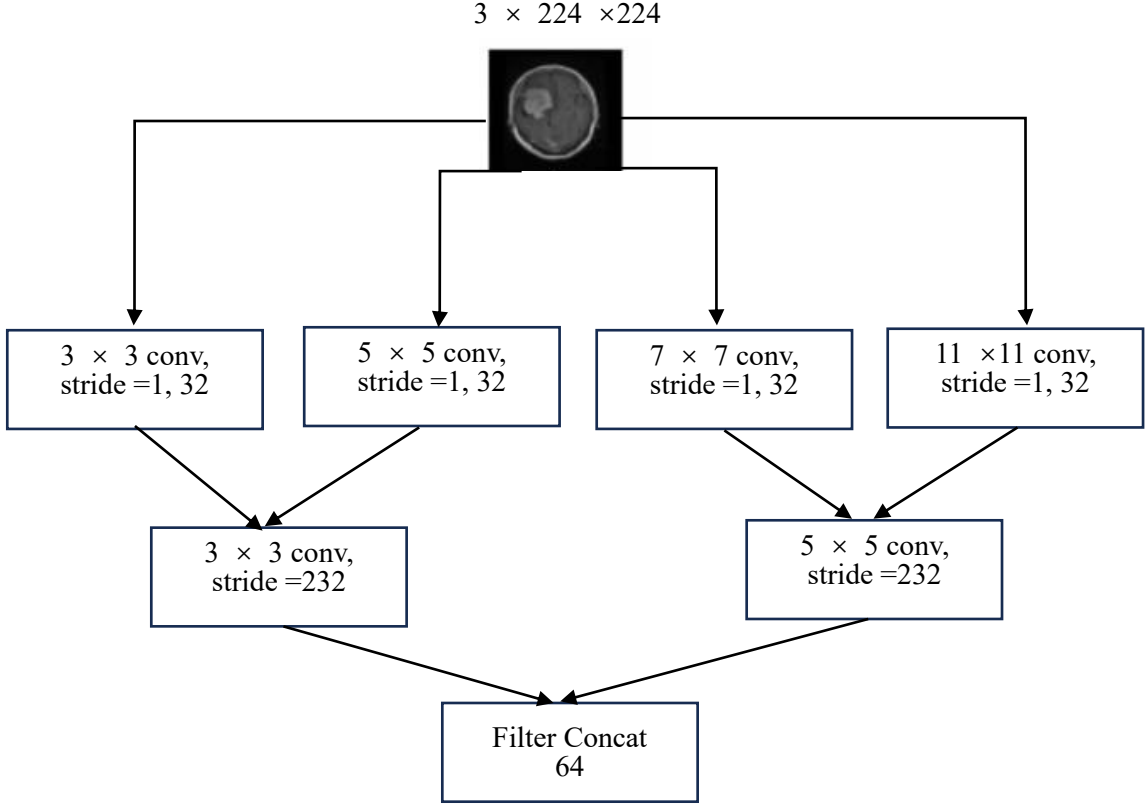

Figure 5 Multi-scale input module

## 2.5 Channel Attention Mechanism

During convolution and pooling, it is assumed that each channel of the feature map has the same importance. However, in real-world problems, the importance of features in different channels may differ. To more accurately capture the differences between channels in the feature map, this project introduces the SE (Squeeze-and-Excitation) channel attention mechanism[16]. It uses a weight matrix to assign different weights to different channels of the image from a channel domain perspective, highlighting more important feature information. SE mainly consists of three parts: squeezing, excitation, and scaling.

The squeezing part uses global average pooling (GAP) to compress the size of the input feature map from $H×W×C$ to $1×1×C$. This involves averaging all pixels in each channel of the feature map to obtain a unique value, as shown in the following formula:

$$Z_c = F_{sq}(u_c) = \frac{1}{H \times W} \sum_{i=1}^{H} \sum_{j=1}^{W} u_c(i,j) \qquad (2)$$

In the formula, $Z_c$ represents the global compressed feature of the $c$-th channel of the input feature map, $F_{sq}$ represents the compression process, $u_c$ represents the feature map of the $c$-th channel of the input feature map, and $H$ and $W$ represent the height and width of the input feature map, respectively. The activation part processes the set obtained from the compression part through two fully connected layers and one activation layer to obtain the desired weight values s, where different values in s represent the weight information of different channels. The formula is as follows:

$$s = F_{ex}(z, W) = \sigma[W_2 \delta(W_1 z)] \quad (3)$$

In the formula, $F_{ex}$ represents the activation process; $z$ represents the output obtained from the compression part; $W$ represents the fully connected layer, $W_1$ and $W_2$ represent the first and second fully connected layers respectively, $\delta$ represents the ReLU activation layer, and $\sigma$ represents the sigmoid activation function.

The reset part multiplies the weight value s of each channel obtained from the activation part with the input before the compression step for the corresponding channel, so as to assign different weights to different channels. The formula is as follows:

$$\tilde{X}_c = F_{scale}(u_c, S_c) = S_c u_c \quad (4)$$

In the formula, $\tilde{X}_c = [\tilde{x}_1, \tilde{x}_2, \tilde{x}_3, ... \tilde{x}_c]$, $\tilde{x} = s_1 u_1$, $F_{scale}$ represents the reset process, $u_c$ is the input feature, and $S_c$ represents the output weights obtained after the input feature $u_c$ undergoes compression and activation processes.

# 3. Experiment and Analysis

## 3.1 Experimental Environment and Parameter Settings

This experiment uses the PyTorch deep learning model framework, with Python 3 as the programming language and PyCharm as the compiler. The operating system is Windows 11, the CPU is an Intel i7-12500H, and the GPU is an NVIDIA RTX 3090 to utilize the powerful graphics processing unit for computation. During network training, the number of epochs was set to 300, and the batch size for each epoch was set to 64. The Adam optimization algorithm was used, with an initial learning rate of 0.001. A linear decay strategy for the learning rate was adopted, meaning the learning rate was reduced to 0.1 times its original value every 100 epochs to accelerate model convergence and alleviate oscillations and overfitting.

## 3.2 Dataset and Preprocessing

The brain tumor MRI dataset used in this experiment comes from 3064 brain tumor MRI images[17] from the FigShare website and 1436 tumor-free MRI images[18] from the PaperwithCode website. The 3064 brain tumor images from FigShare represent 233 patients with three types of brain tumors: meningioma (708 slices), glioma (1426 slices), and pituitary adenoma (930 slices). The tumor type in each slice was labeled by the radiologist. This dataset is one of the most widely used public datasets for current brain tumor classification tasks. It is provided in Matlab format, with each file storing a structure containing image information, such as the relationship between tumor type and ground truth labels (1 for meningioma, 2 for glioma, 3 for pituitary adenoma). Figure 6 shows the different types of brain tumors and tumor-free MRI images after converting the Matlab format files to image files.

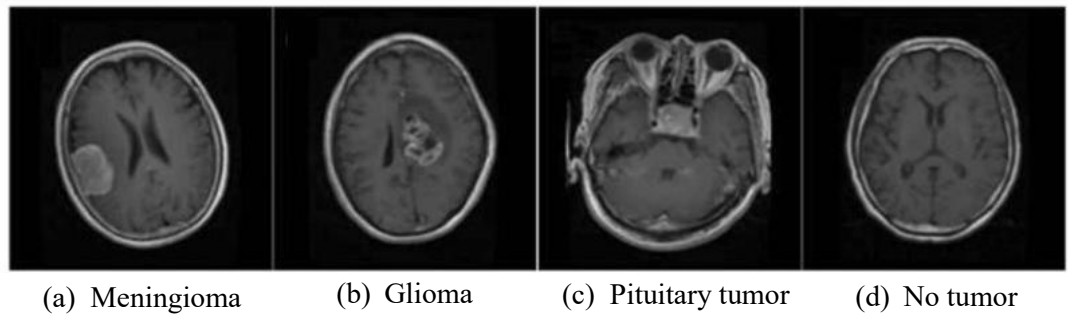

(a) Meningioma    (b) Glioma    (c) Pituitary tumor    (d) No tumor

Figure 6 Images of tumors and tumor-free areas

After converting the entire dataset into image files, one-fifth of them were randomly selected as the test set. The remaining dataset was then randomly divided into training and validation sets in a 4:1 ratio, as detailed in Table 1. Before feeding the dataset into the training network, preprocessing was required. First, each image was randomly cropped and resized to a fixed 224×224 size for easy input into the network. Next, the cropped images were standardized to eliminate scale differences and improve the stability of gradient descent. Due to the relatively small size of the dataset, data augmentation was employed to increase its diversity. The input images were randomly flipped horizontally, vertically, and brightened to ensure that the data input to the network was not identical in each training cycle. This mitigated potential overfitting during model training and improved the model's generalization performance.

Table 1 Dataset distribution

| Tumor types | Training set | Validation set | Test set | total |
|---|---|---|---|---|
| **Meningioma** | 454 | 113 | 141 | 708 |
| **Glioma** | 913 | 228 | 285 | 1426 |
| **Pituitary** | 596 | 148 | 186 | 930 |
| **No tumor** | 920 | 229 | 287 | 1436 |

## 3.3 Evaluation Metrics

Evaluation metrics are used to measure the quality of model training. This paper adopts the

following evaluation metrics as the standard for experimental results:

1) Accuracy: Represents the proportion of samples in the test results where the predicted value is the same as the true value, out of the total sample. The formula is as follows:

$$Accuracy = \frac{TP + TN}{TP + FP + TN + FN} \quad (6)$$

2) Precision: Represents the percentage of true positive samples out of all samples predicted to be positive in the test results. The formula is as follows:

$$Precision = \frac{TP}{TP+FP} \quad (7)$$

3) Recall: Represents the percentage of samples where the true value is positive, and also the percentage of samples where the predicted value is positive. The formula is as follows:

$$Recall = \frac{TP}{TP+FN} \quad (8)$$

4) Specificity: Represents the proportion of samples with negative predicted values (also negative) among samples with negative true values in the test results. The formula is as follows:

$$Specificity = \frac{TN}{FP+TN} \quad (9)$$

5) $F_{1\ score}$: Represents a weighted average of precision and recall. A higher $F_{1\ score}$ indicates a more robust model. The formula is as follows:

$$F_{1\ score} = 2 \times \frac{Precision \times Recall}{Precision+Recall} \quad (10)$$

In the above formulas, TP represents the number of positive class samples correctly classified as positive, i.e., the number of positive class samples successfully identified by the model; TN represents the number of negative class samples correctly classified as negative, i.e., the number of negative class samples successfully identified by the model; FP represents the number of negative class samples incorrectly classified as positive, i.e., the number of negative class samples incorrectly labeled as positive; and FN represents the number of positive class samples incorrectly classified as negative, i.e., the number of positive class samples incorrectly labeled as negative.

## 3.4 Experimental Results

To further investigate the effectiveness of the proposed method and the impact of the improved modules on the accuracy of the original ResNet34 network, this paper combines the ResNet34 network with the Inception v2 module, the multi-scale input module, and the attention mechanism module, respectively. An ablation study was used to verify the integration of these modules into the ResNet34 network, and the network was trained on the training dataset. The performance of these five models was evaluated on the test dataset, and the average test results are shown in Table 2. Table 3 lists other evaluation metrics for each model. This experimental design aims to comprehensively evaluate the impact of different modules on model performance, providing profound insights for model optimization.

Table 2 Model's average test accuracy

| Model | ResNet34 | A | B | C | Accuracy/% |
|---|---|---|---|---|---|
| 1 | √ | × | × | × | 97.71 |
| 2 | √ | √ | × | × | 98.33 |
| 3 | √ | × | √ | × | 98.22 |
| 4 | √ | × | × | √ | 98.28 |
| 5 | √ | √ | √ | √ | 98.82 |

In Table 2, A represents the Inception v2 module, B represents the multi-scale input module, and C represents the SE attention mechanism module. Model 1 represents a ResNet34 network only; Model 2 represents a new network formed by replacing the downsampling layer in ResNet34 with the Inception v2 module; Model 3 represents a new network formed by combining ResNet34 with the multi-scale input module; Model 4 represents a new network formed by adding the SE attention mechanism module to ResNet34; and Model 5 represents a new network formed by incorporating modules A, B, and C (a total of three modules) into ResNet34. The average accuracy results of these five different network architectures on the test set show that the improved networks with added modules all have higher accuracy in brain tumor image classification than the original networks, indicating that these modules improve the performance of the original model in brain tumor image classification. Among them, the network model integrating three modules has the best accuracy, reaching 98.82%, which is about 1.1 percentage points higher than the ResNet34 network.

Table 3 Average evaluation index of each model    %

| Model | Type | Precision | Recall | Specificity | $F_1$ |
|---|---|---|---|---|---|
| 1 | Glioma | 98.18 | 98.25 | 99.15 | 98.21 |
|  | Meningioma | 94.06 | 93.9 | 98.89 | 93.97 |
|  | No tumor | 100 | 99.3 | 100 | 99.65 |
|  | Pituitary | 96.28 | 97.31 | 99.02 | 96.79 |
|  | Average | 97.13 | 97.19 | 99.27 | 97.16 |
| 2 | Glioma | 97.92 | 99.12 | 99.02 | 98.52 |

| | Type | Precision | Recall | Specificity | F1 |
|---|---|---|---|---|---|
| | Meningioma | 96.08 | 95.74 | 99.27 | 95.91 |
| | No tumor | 100 | 99.48 | 100 | 99.74 |
| | Pituitary | 98.1 | 97.31 | 99.51 | 97.71 |
| | Average | 98.03 | 97.91 | 99.45 | 97.97 |
| 3 | Glioma | 98.94 | 98.42 | 99.51 | 98.68 |
| | Meningioma | 94.41 | 95.75 | 98.95 | 95.07 |
| | No tumor | 100 | 99.83 | 100 | 99.91 |
| | Pituitary | 97.31 | 97.31 | 99.3 | 97.31 |
| | Average | 97.67 | 97.83 | 99.44 | 97.74 |
| 4 | Glioma | 98. 44 | 99.12 | 99.1 | 98.6 |
| | Meningioma | 96.09 | 95.75 | 99.27 | 95.92 |
| | No tumor | 100 | 98.95 | 100 | 99.47 |
| | Pituitary | 97.59 | 97.85 | 99.37 | 97.72 |
| | Average | 98.03 | 97.92 | 99.44 | 97.93 |
| 5 | Glioma | 97.93 | 98.88 | 99.74 | 99.16 |
| | Meningioma | 96.12 | 97.87 | 99.26 | 96.98 |
| | No tumor | 100 | 99.79 | 100 | 99.9 |
| | Pituitary | 97.95 | 97.74 | 99.47 | 97.85 |
| | Average | 98.33 | 98.57 | 99.62 | 98.47 |

In Table 3, under the category of "type," glioma, meningioma, no tumor, and pituitary adenoma represent glioma, meningioma, no tumor, and pituitary adenoma, respectively, and "average" represents the average value of the three. Based on the detailed data analysis in Tables 2 and 3, it can be clearly observed that models 2, 3, 4, and 5 outperform model 1 in multiple performance metrics, including precision, recall, specificity, and F1 score. This result demonstrates the performance improvement of the improved network models; compared to the original network, the improved models exhibit higher levels of performance across all evaluation metrics.

To ensure the reliability and robustness of the model, a five-fold cross-validation method was used for extensive experimental verification. Figure 7 compares the training loss and validation accuracy of one of the improved models with the original ResNet34 model. The improved network model has a lower training loss and a higher validation accuracy than the original model, indicating that the training results of the improved model are closer to the true values. The confusion matrix of the test results helps to understand the classification results and performance of the model in more detail, so the confusion matrix after each validation was calculated, as shown in Figure 8. The horizontal axis represents the true value, and the vertical axis represents the predicted value. It can be seen that only a very small number of brain tumor MRI images in each class were incorrectly predicted as other classes, and the vast majority of brain tumor MRI images were correctly classified, with the prediction of no tumors being almost 100%. The confusion matrix can also be used to calculate the model's performance on multiple evaluation indicators, as detailed in Table 4.

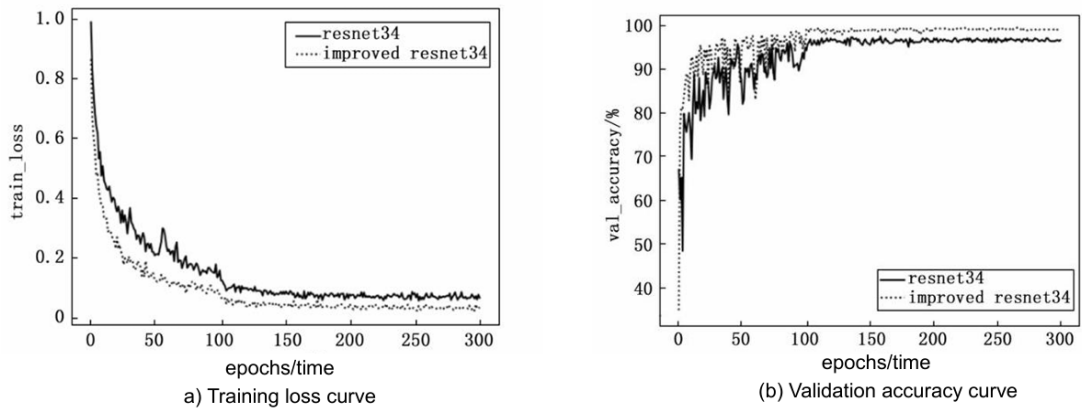

Figure 7 Training loss and validation accuracy curves of the post-processed network and the ResNet34 network

Table 4　The average evaluation metric of the five-fold cross-validation of the improved network model　%

| K-Fold | Accuracy | Precision | Recall | Specificity | $F_1$ |
|---|---|---|---|---|---|
| 1 | 99 | 98.64 | 98.84 | 99.67 | 98.74 |
| 2 | 98.67 | 98.15 | 98.58 | 99.59 | 98.35 |
| 3 | 98.89 | 98.58 | 98.62 | 99.65 | 98.6 |
| 4 | 98.67 | 98.17 | 98.44 | 99.59 | 98.3 |
| 5 | 98.89 | 98.54 | 98.62 | 99.65 | 98.58 |
| Average | 98.82 | 98.42 | 98.62 | 99.63 | 98.51 |

A thorough analysis of the data in Table 4 revealed that the improved model exhibited similar test accuracy across five cross-validations, demonstrating its performance stability. In addition to accuracy, key performance indicators such as precision, recall, specificity, and F-score were comprehensively evaluated. These indicators also showed similar results across the five validations, further confirming the reliability and robustness of the improved model. To more comprehensively evaluate the model's performance, a detailed comparison was made with several widely used classification algorithms in current academic and practical fields. All models were evaluated on the same test set, and the detailed results are presented in Table 5.

Furthermore, this paper provides a detailed statistical analysis of the parameter counts of the new model and the ResNet34 model. The ResNet34 network model has approximately 21.3M parameters, while the proposed new model has approximately 17.3M parameters, about 80% of the ResNet34 parameter count. This indicates that the new model maintains high performance while reducing space complexity, providing feasibility for resource efficiency and model lightweighting. A comparison of the proposed new model with relevant research in the field of brain tumor classification in recent years is shown in Table 6; our model exhibits relatively higher accuracy.

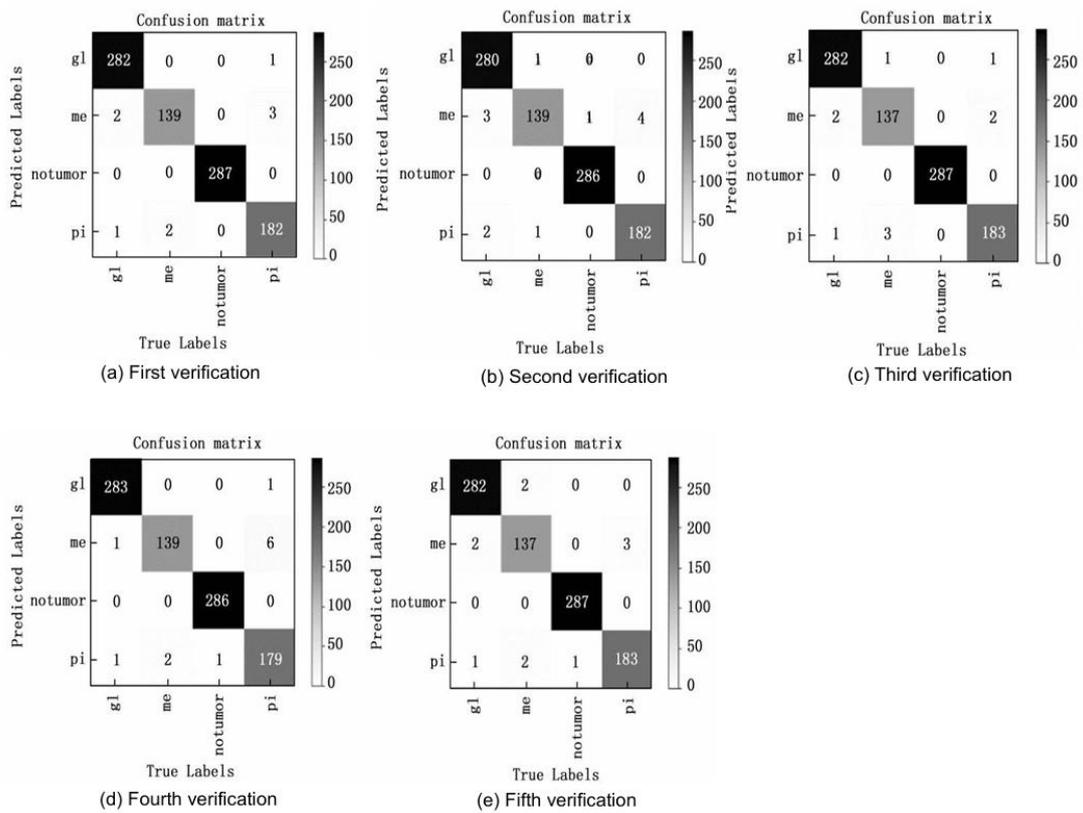

Figure 8    The prediction confusion matrix of the post-processing network model with 5-fold cross-validation

Table 5    The average evaluation index of other classification networks after improvement

| K-Fold | Accuracy | Precision | Recall | Specificity | $F_1$ |
|---|---|---|---|---|---|
| **Vgg16** | 95.77 | 95.59 | 94.41 | 98.55 | 94.93 |
| **ResNet50** | 97.88 | 97.96 | 96.99 | 99.27 | 97.4 |
| **ResNext101** | 98.22 | 97.79 | 97.83 | 99.43 | 97.8 |
| GoogleNet | 97.66 | 97.24 | 96.66 | 99.23 | 97.1 |
| EfficientNet V2 | 98.22 | 97.9 | 97.56 | 99.41 | 97.72 |
| DenseNet | 98.22 | 98.06 | 97.69 | 99.4 | 97.87 |
| **Ours** | 98.82 | 98.42 | 98.62 | 99.63 | 98.51 |

Table 6　　Accuracy comparison of relevant studies in recent years

| Author | Year | Approach | Accuracy/ % |
|---|---|---|---|
| **Swati et alt** | 2019 | VGG19 + Fine Tuning | 94. 82 |
| **Kaplan et al.** | 2020 | LBP | 95.56 |
| **Kumar et al[** | 2021 | ResNet-50 +Global Average Pooling | 97.48 |
| **Huang et al** | 2022 | Multiscale Residual Network | 98.58 |
| **ours** | — | ResNet34+Multiscale Inputs + Inception v2+ SE | 98.82 |

# 4. Conclusion

Traditional manual classification of brain tumor medical images is time-consuming and labor-intensive, shallow CNN models have low classification accuracy, and stacked deep networks are prone to gradient vanishing. This paper combines the ResNet34 residual network model with multi-scale concepts and an attention mechanism module for a four-class classification task of brain tumor MRI images. By leveraging the powerful ability of residual networks to mitigate gradient vanishing with depth in deep neural networks, convolutional neural networks can be stacked to a certain depth. To address the information loss that may result from ResNet34's single-scale feature extraction and downsampling of residual blocks, this paper introduces a multi-scale input and an incidence v2 module to ensure diverse feature extraction while avoiding information loss. Furthermore, an SE attention mechanism module is added to assign different weights to different channels of the image from a channel domain perspective, thereby more effectively capturing key feature information. Finally, experiments using five-fold cross-validation show that the improved model achieves an average accuracy of 98.82%, which is approximately 1.1% higher than the ResNet34 model, and the network model has fewer parameters. Compared to other classification algorithms and related research in recent years, this model also has a relatively higher accuracy.

The specific process and implementation steps of applying the improved model to brain tumor image classification are as follows: 1) First, divide the brain tumor image dataset into a training set, a validation set, and a test set; 2) Perform preprocessing operations such as cropping, standardization, and transformation on the dataset; 3) Train the processed training set using the improved network model through the training module to generate training weights; 4) Test each training weight on the validation set and retain the optimal training weights by verifying the accuracy; 5) After obtaining the optimal training weights, test them on the test set. If their performance on the test set is also good and relatively stable, the training weights and test code module can be combined and deployed into diagnostic software or healthcare systems to detect brain tumor types. This will help doctors diagnose brain tumor types more accurately and improve diagnostic efficiency and accuracy.